\documentclass[10pt, a4paper]{article}
\usepackage{lrec2022} 
\usepackage{multibib}
\newcites{languageresource}{Language Resources}
\usepackage{graphicx}
\usepackage{tabularx}
\usepackage{soul}
\usepackage{balance}
\usepackage{titlesec}
\titleformat{\section}{\normalfont\large\bfseries\center}{\thesection.}{1em}{}
\titleformat{\subsection}{\normalfont\SmallTitleFont\bfseries\raggedright}{\thesubsection.}{1em}{}
\titleformat{\subsubsection}{\normalfont\normalsize\bfseries\raggedright}{\thesubsubsection.}{1em}{}
\renewcommand\thesection{\arabic{section}}
\renewcommand\thesubsection{\thesection.\arabic{subsection}}
\renewcommand\thesubsubsection{\thesubsection.\arabic{subsubsection}}

\usepackage{epstopdf}
\usepackage[utf8]{inputenc}

\usepackage{hyperref}
\usepackage{xstring}

\usepackage{color}

\title{Conversational Analysis of Daily Dialog Data using\\Polite Emotional Dialogue Acts}

\name{Chandrakant Bothe$^\dagger$ and Stefan Wemter$^\ddagger$} 

\address{
        $^\dagger$Foviatech GmbH\\
        \url{chandrakant.bothe@foviatech.com}\\
        $^\ddagger$Knowledge Technology, Department of Informatics, University of Hamburg\\ 
        \url{stefan.wermter@uni-hamburg.de}\\
        }

\abstract{
Many socio-linguistic cues are used in conversational analysis, such as emotion, sentiment, and dialogue acts.
One of the fundamental cues is politeness, which linguistically possesses properties such as social manners useful in conversational analysis.
This article presents findings of polite emotional dialogue act associations, where we can correlate the relationships between the socio-linguistic cues.
We confirm our hypothesis that the utterances with the emotion classes Anger and Disgust are more likely to be impolite. 
At the same time, Happiness and Sadness are more likely to be polite.
A less expectable phenomenon occurs with dialogue acts Inform and Commissive which contain more polite utterances than Question and Directive.
Finally, we conclude on the future work of these findings to extend the learning of social behaviours using politeness.
\\ \newline 
 }

\begin{document}

\maketitleabstract

\section{Introduction}

Conversational analysis can potentially be enhanced with socio-linguistic politeness cues along with other linguistic cues \cite{kasper1990linguistic,russell1999core,P13dan2013compPoliteness}.
Literature suggests that analysis and use of social cues is beneficial for human-robot interaction
\cite{salem2014marhaba,BARROS2015140,castro2016effects,bothe2020conversational}.
Our article focuses on an experiment by primarily adding linguistic politeness cues, expressed using polite phrases, to conversational analysis along with emotional states and dialogue acts where we aim to discover correlations between these cues.

\begin{figure}[t!]
\begin{center}
\includegraphics[width=0.5\textwidth]{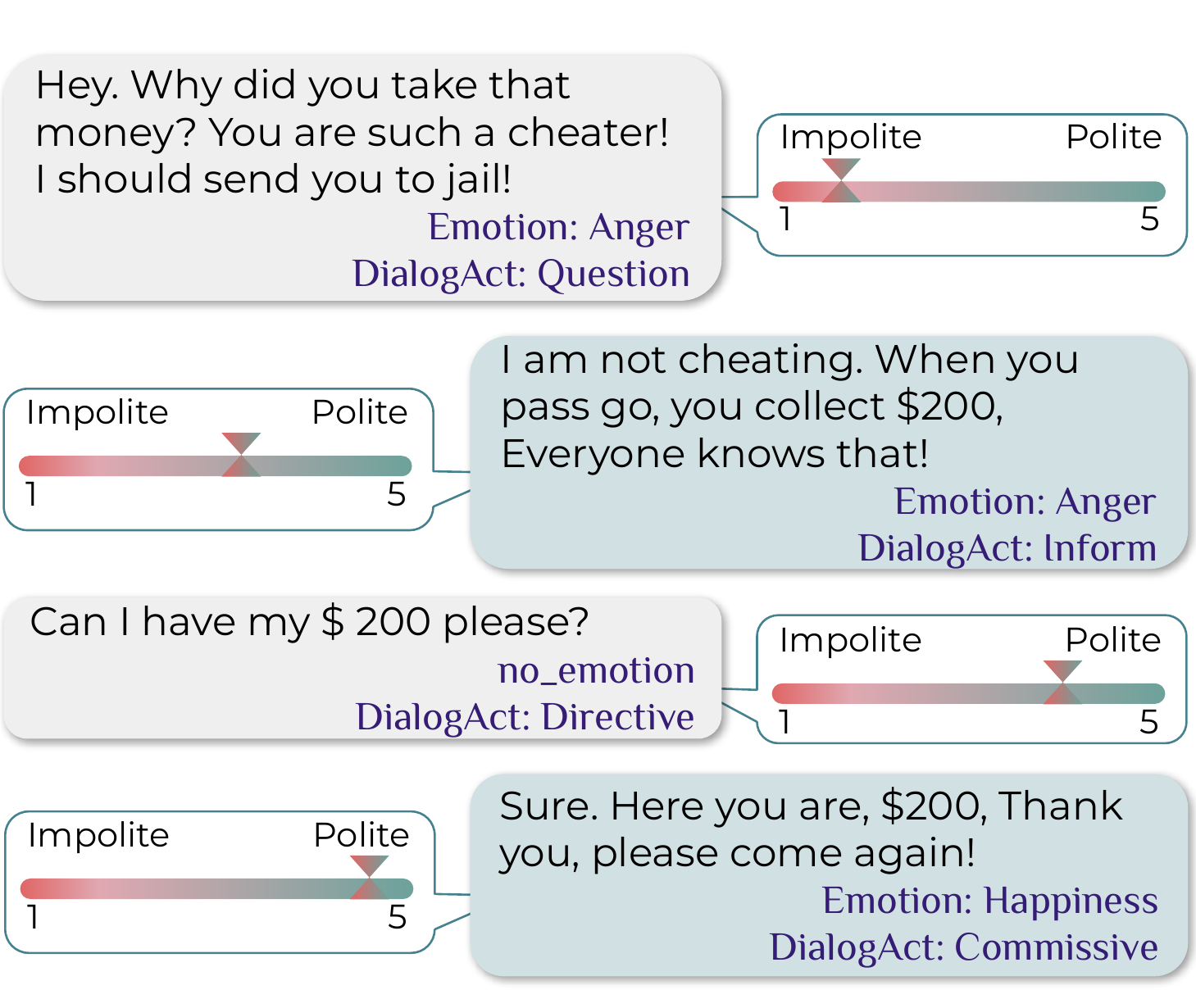} 
\caption{Polite Emotional Dialogue Act labels from the DailyDialog dataset showing the utterances with emotion Anger is very Impolite while with emotion Happiness is very Polite in the bi-turn dialogue flows (Emotion and DialogAct labels are from the already annotated DailyDialog dataset)}
\label{ang-happ-pols}
\end{center}
\end{figure}

For our experiment, we will explore DailyDialog dataset \cite{dailydialog2017}, a multi-turn dialogue dataset, which is pre-annotated with emotion and dialogue act labels.
Furthermore, the dataset will be augmented with politeness labels ranging from 1 to 5 by leveraging a pre-trained model \cite{jiajun2021conversation}. 
By analyzing the correlation between politeness and emotional states, we discover that utterances with certain emotion classes are more polite or impolite than others. 
At least frequently, it appears natural to use polite utterances in a happy emotional state, whereas to use impolite utterances in an angry state.
This phenomenon is precisely the motivation behind this experiment and we can find such utterance examples illustrated in Figure~\ref{ang-happ-pols}.
The conversation in this example shows how an impolite utterance is used in the Anger emotion state and then shifts with a polite utterance to the Happiness emotion state.
The results show this phenomenon statistically occurring significantly in the given dataset.

The conversational behaviours learned with these cues might be beneficial to drive the dialogue flow in complex polite human-robot interaction setup, which was not possible in a straightforward dialogue-based navigation system \cite{bothe2018towards}.
Additionally, the results could also be applied in pragmatic conversational analysis where politeness strategies could assist in dialogue flow \cite{jiajun2021conversation} and theoretically analyse the core affect study \cite{russell1999core}.
The politeness annotated data and the results are available in the GitHub repository \textit{bothe/politeEDAs}\footnote{\url{https://github.com/bothe/politeEDAs}}.

\begin{table*}[t!]
\centering
\begin{tabular}{lrrrrrrr}
\hline      \multicolumn{4}{c}{Emotions} & \multicolumn{4}{c}{Dialogue Acts} \\ 
\hline      & Utterances   & \% total & \% Emotion & Inform & Question & Directive & Commissive \\ \hline
Anger       & 1022  & 1.00  &  5.87 &   615 &   174 &   132 &  101 \\
Disgust     & 353   & 0.34  &  2.03 &   291 &    30 &    18 &   14 \\
Fear        & 174   & 0.17  &  1.00 &   106 &    27 &    21 &   20 \\
Happiness   & 12885 & 12.61 & 74.02 &  7830 &  2158 &  1476 & 1421 \\
Sadness     & 1150  & 1.13  &  6.61 &   809 &   190 &    95 &   56 \\
Surprise    & 1823  & 1.78  & 10.47 &  1122 &   565 &    72 &   64 \\
no\_emotion & 85572 & 83.78 &    -- & 36316 & 26549 & 15542 & 7165 \\
\hline
Total       &102979 &  --   &    -- & 47089 & 29693 & 17356 & 8841 \\
\hline
\end{tabular}
\caption{\label{emo-data-table}Statistics of the number of utterances in Emotion and Dialogue Act classes in the DailyDialog dataset}
\end{table*}

\section{Related Work}

Perceiving emotions in conversation provides affective information of the conversation partners; similarly, perceiving politeness in conversation provides cues about their social manners/behaviours.
The emotion and dialogue act relationships are presented by \cite{bothe2019enriching} for Emotional Dialogue Acts using previous work on emotion and dialogue act recognition \cite{labothe2017WASSA2017,bothe2018discourse}. 
For example, Accept/Agree and Thanking dialogue acts often occur with the Joy emotion, Apology with Sadness, Reject with Anger, and Acknowledgements with Surprise.
Similarly, the relationship of politeness and emotion are evidently discussed much in the literature as some of the most critical social cues \cite{Langlotz2017polemo,renner2020directness,Culpeper2021politeness}.
A computational linguistic study shows how machines learn politeness, for example, the words \textit{please} and \textit{could you} signal on the heatmaps of sentences \cite{aubakirova2016interpoliteness}.
Contrary, prosodic information provides an additional dimension to politeness as the exact phrase might be uttered differently \cite{culpeper2011s}.
In fact, \cite{brown1987politeness} mention the display of emotions or lack of control of emotions as positive politeness strategies or potentially face-threatening acts of politeness.

Thus, this study presents an analysis of linguistic politeness by understanding how ``appropriate levels of affect" are conveyed in the conversational interaction \cite{holmes2015power,Langlotz2017polemo,kumar2021exploratory}.
In human-robot interaction, politeness as a social cue plays a vital role to foster socially engaging interaction with robots \cite{steinfeld2006common,srinivasan2016help,bothe2018towards}.
However, this article explores relations between the socio-linguistic cues, finding out how polite the utterances are against their respective emotion and dialogue act classes.

\section{Approach}

Our goal is to analyze the socio-linguistic cues to find meaningful correlations between them.
First, we explore the dataset, DailyDialog, which is pre-annotated with emotion and dialogue act labels.
Second, we annotate the utterances with politeness values using a politeness analyzer, more precisely each utterance with a degree of politeness.
Finally, we analyze the annotated utterances with politeness against the emotion and dialogue act classes in the dataset to identify their associations.
Eventually, we conclude with the results and findings and provide future conversational analysis work to extend this experiment.

\begin{figure}[b!]
\begin{center}
\includegraphics[width=0.5\textwidth]{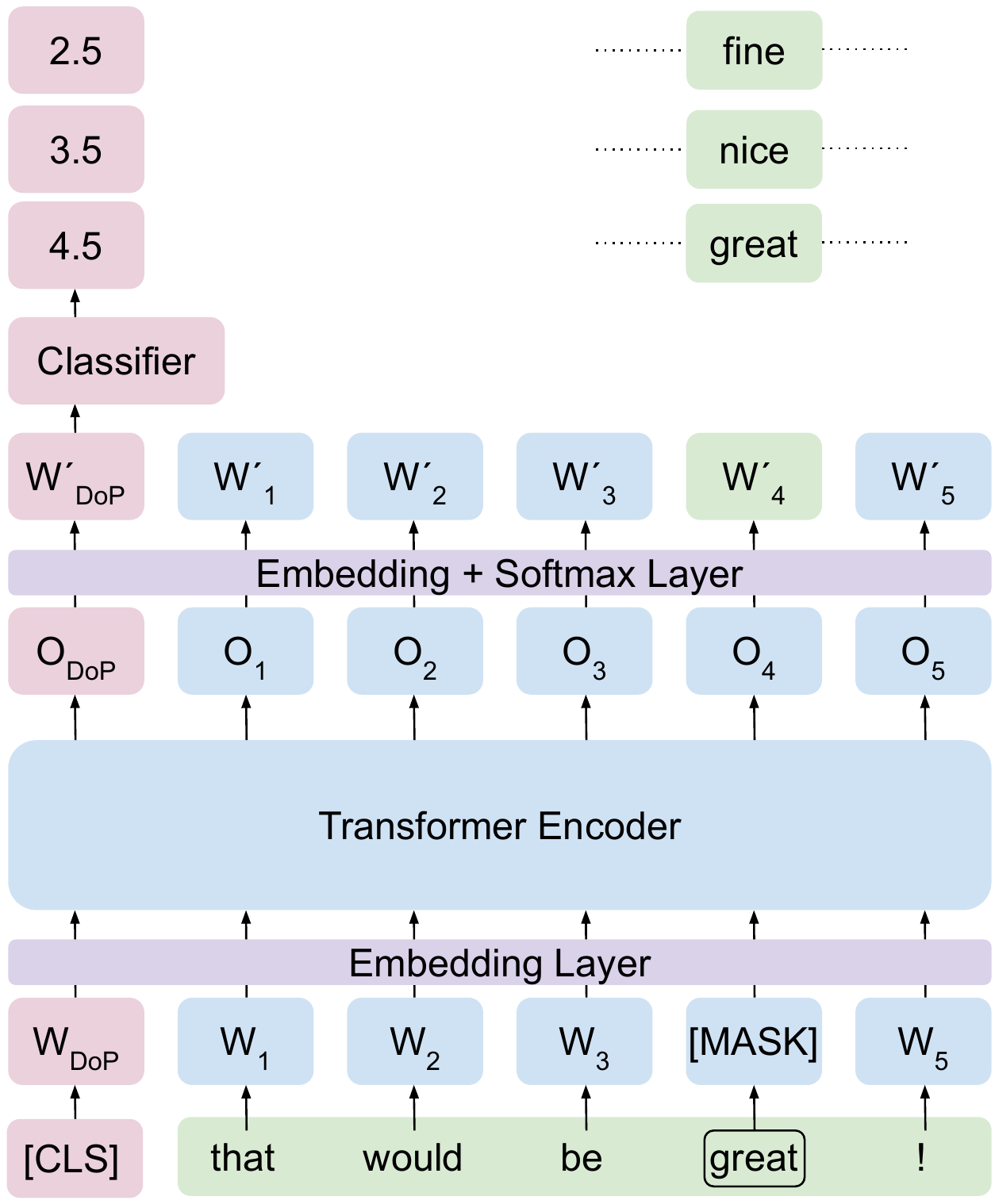} 
\caption{BERT-based masked language model in a classification setup for degree-of-politeness (DoP) recognition on the scale of range between 1 and 5 where a lower value indicates impolite while higher indicates polite whereas around 3 indicates neutral}
\label{fig:bert}
\end{center}
\end{figure}

\begin{table*}[t!]
\centering
\begin{tabular}{llll}
\hline DoP & Dialogue Act & Emotion & Utterance  \\ \hline 

4.63 & Inform & Happiness & OK! Thank you very much! \\
4.63 & Inform & Happiness & Sure, that would be great! Thank you!\\
4.63 & Commissive & Happiness & That would be great! Thanks a lot!\\
4.62 & Inform & Happiness & Thank you, thank you, thanks again.\\
4.62 & Inform & Happiness & This is exciting! Thank you so much!\\
 \hline
 
3.67 & Question & no\_emotion & That's right. What do you think we should do as a hobby? \\
3.67 & Inform & no\_emotion & Mr Jurgen, yes, the remittance has been successful. \\
3.67 & Inform & no\_emotion & Yes. We all loved the celebration of our city's birthday.\\
3.67 & Inform & no\_emotion & Ok, I’ll pay by card then.\\
3.67 & Commissive & no\_emotion & Ok. I won't get into trouble. \\
\hline


3.34 & Inform & Happiness & I went to the tutoring service centre on campus today and got a job. \\
3.34 & Inform & Happiness & I never knew there were so many fun things to do on a farm. \\
3.34 & Inform & no\_emotion & Just a minute. It's ten to nine by my watch. \\
3.34 & Inform & no\_emotion & You may check out books or videos. \\
3.34 & Question &  no\_emotion & That’s a small fee? \\
\hline

2.95 & Commissive & Anger & Mike, you're late again. \\
2.95 & Inform & no\_emotion & Ha! It's not like you've ever been one to beat around the bush. \\
2.94 & Inform & no\_emotion & My rear bumper is messed up. \\
2.94 & Inform & no\_emotion & No, they won't. They are shrink proof.\\
2.94 & Question & no\_emotion & Why don't you wear a scarf? \\
\hline

1.59 & Inform & Disgust & Don’t dress like that. You’ll make fool yourself. \\ 
1.51 & Commissive & Anger & Make it work, Geoff. You would say that, wouldn’t you. \\
1.47 & Directive & no\_emotion & Get up, you lazybones! \\
1.46 & Directive & Surprise & You idiot! Don’t say that! Do you want this job, or not? \\
1.37 & Directive & Anger & Get out of my store, you jerk! \\
\hline
\end{tabular}
\caption{\label{anno-pols-table} Examples from the DailyDialog dataset showing five utterances in the selected ranges of DoP (degree of politeness) with highest values (first block), middle values (three blocks) and lowest values (last block) annotated by the politeness analyzer, which provides a sense of the annotated utterances with different levels of politeness}
\end{table*}


\subsection{DailyDialog Dataset}

The DailyDialog dataset contains conversation topics of daily life such as ordinary life and financial topics.
It contains bi-turn dialogue flows like Question-Inform and Directive-Commissive dialogue acts.
Thus, the dataset is annotated with those four fundamental dialogue acts to follow unique multi-turn dialogue flow patterns \cite{dailydialog2017}. 
This dataset is manually labelled with six emotion classes and a no\_emotion class.
The statistics of the dataset is presented in Table~\ref{emo-data-table}, we can see that the no\_emotion class dominates the total number of utterances.
However, Happiness dominates within the emotion classes, whereas Fear contains the least number of utterances.
Further, Table~\ref{emo-data-table} also presents the number of utterances for the four dialogue act classes in the dataset for their corresponding emotion classes.

\subsection{Politeness Analyzer}

To annotate the utterances, we will use a politeness analyzer from the recent work of \cite{jiajun2021conversation}, where they combine two datasets from \cite{P13dan2013compPoliteness} and \cite{wang2018convosupport}.
The politeness regressor model is obtained on these datasets by pre-training a BERT-based model \cite{devlin2018bert} on Reddit data using masked language modelling, as shown in Figure~\ref{fig:bert}. 
The word embeddings are encoded with the transformer encoder, which later is used to classify the politeness values. 
As the words are masked, various words can be detected given different levels of the degree of politeness (DoP).
Similarly, given different words, the classifier model can determine various levels of DoP. 
This model can also be useful to apply utterance-level context-based politeness learning using hierarchical recurrent neural networks \cite{bothe2018interspeech}.

The encoder is pre-trained on the target dataset and fine-tuned on the above mentioned two politeness datasets. 
The final model is obtained at an average Pearson correlation of 0.66 with human judgments from both the datasets and made available online at the GitHub repository \textit{wujunjie1998/Politenessr}\footnote{\url{https://github.com/wujunjie1998/Politenessr/}}.
The model provides a degree of politeness on a scale between 1 and 5 (presented as float values), where around 3 indicates a neutral state: neither polite nor impolite.
In this way, we achieve a very fine-grained annotation to each utterance in the given dataset.

\begin{figure*}[t!]
\begin{center}
\includegraphics[width=\textwidth]{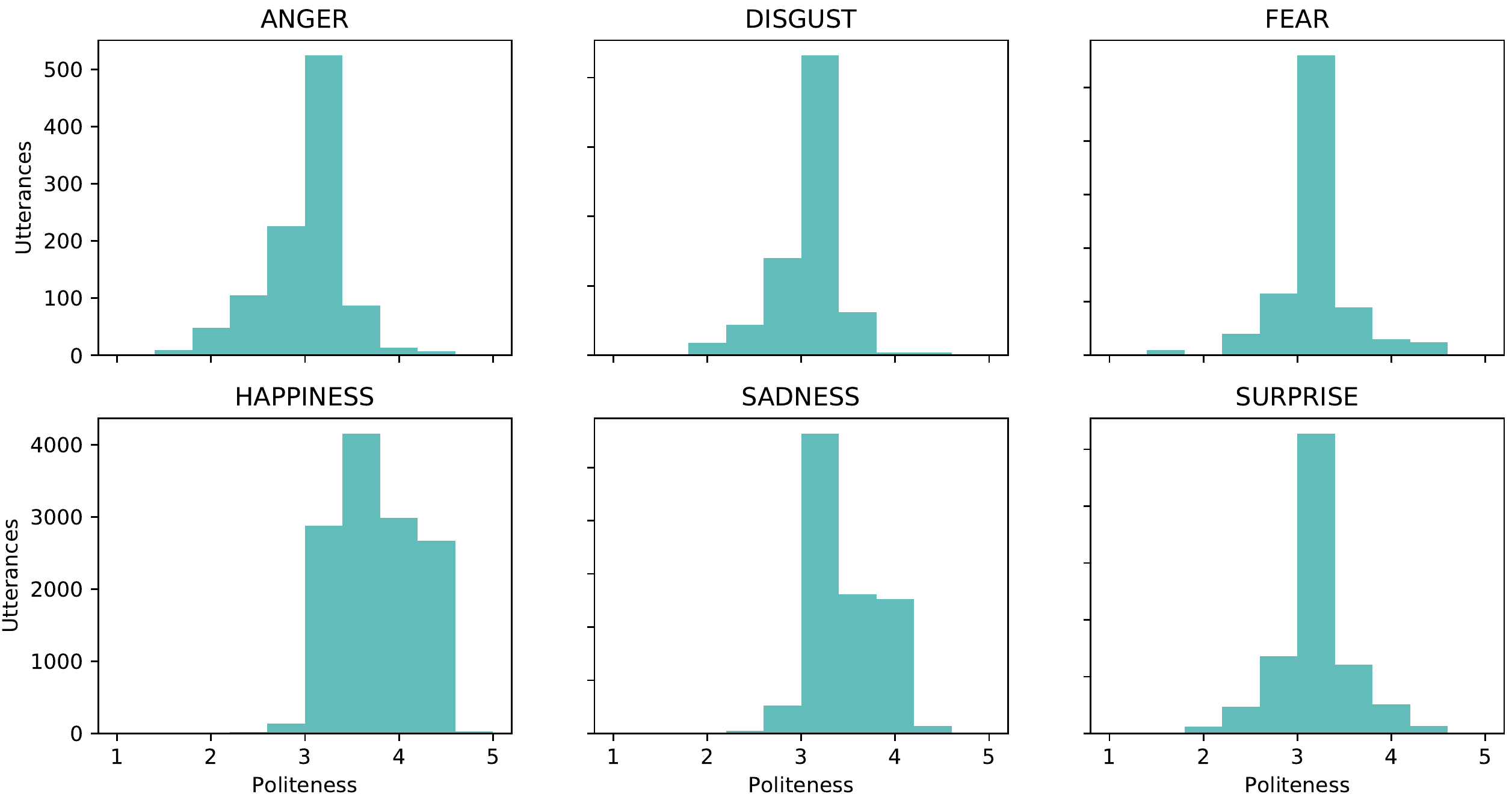} 
\caption{Politeness histograms for the emotion categories in the DailyDialog dataset}
\label{pol_histo_emo}
\end{center}
\end{figure*}

\section{Experiments and Results}

We annotate all utterances in the DailyDialog dataset for politeness values in the range between 1 and 5 using the politeness analyzer.
The annotated utterances are arranged in decreasing order to find the examples of polite, neutral and impolite utterances. 
Then they are sorted according to a scenario, first for every emotion class and second for every dialogue act class. 

All the utterances are arranged in decreasing order, and Table~\ref{anno-pols-table} presents top polite, neutral and impolite utterances from the DailyDialog dataset annotated with the politeness analyzer.
Five examples are presented from the top 10 in each extreme value range (only 5 examples are shown out of the top 10 to eliminate similar or repeated utterances).
The value ranges are chosen as follows: all the utterances are arranged in decreasing order of the politeness values, and then 10 utterances are selected having two extreme values (top very polite and very impolite) and middle three ranges (very polite to centre, centre, and very impolite to centre).
We find that most of the utterances in the first block with Happiness and `thank you' phrases are recognized with a significantly higher degree of politeness. 
In contrast, most of the utterances in the last block with the Anger and Disgust emotion classes containing rude words like `fool', `idiot', or `jerk' are recognized with a significantly lower degree of politeness.
Furthermore, we can see that the second block contains comparatively less polite utterances than the first block.
Similarly, the third block contains comparatively more polite utterances than the last block.
On the other hand, the middle block contains relatively neutral utterances, at least reasonably neutral compared to other blocks.
With these examples, we assure that the politeness analyzer annotates the utterances with significantly well degrees of politeness.

In the next phase, we sort all the utterances for the six emotion classes in the dataset.
Then we plot histograms of the politeness values of the utterances against each emotion category, which is shown in Figure~\ref{pol_histo_emo}.
As we can notice in these histograms, utterances in the emotion classes Anger and Disgust are neutral and impolite on the politeness scale. 
On the other hand, many utterances in the Happiness and Sadness emotion classes are polite. 
Fear and Surprise emotion classes are mostly neutral on the politeness scale compared to other emotion classes.
This analysis gives an exceptional insight into the relationship between emotional states and politeness strategies used in dialogues.

\begin{figure}[b!]
\begin{center}
\includegraphics[width=0.5\textwidth]{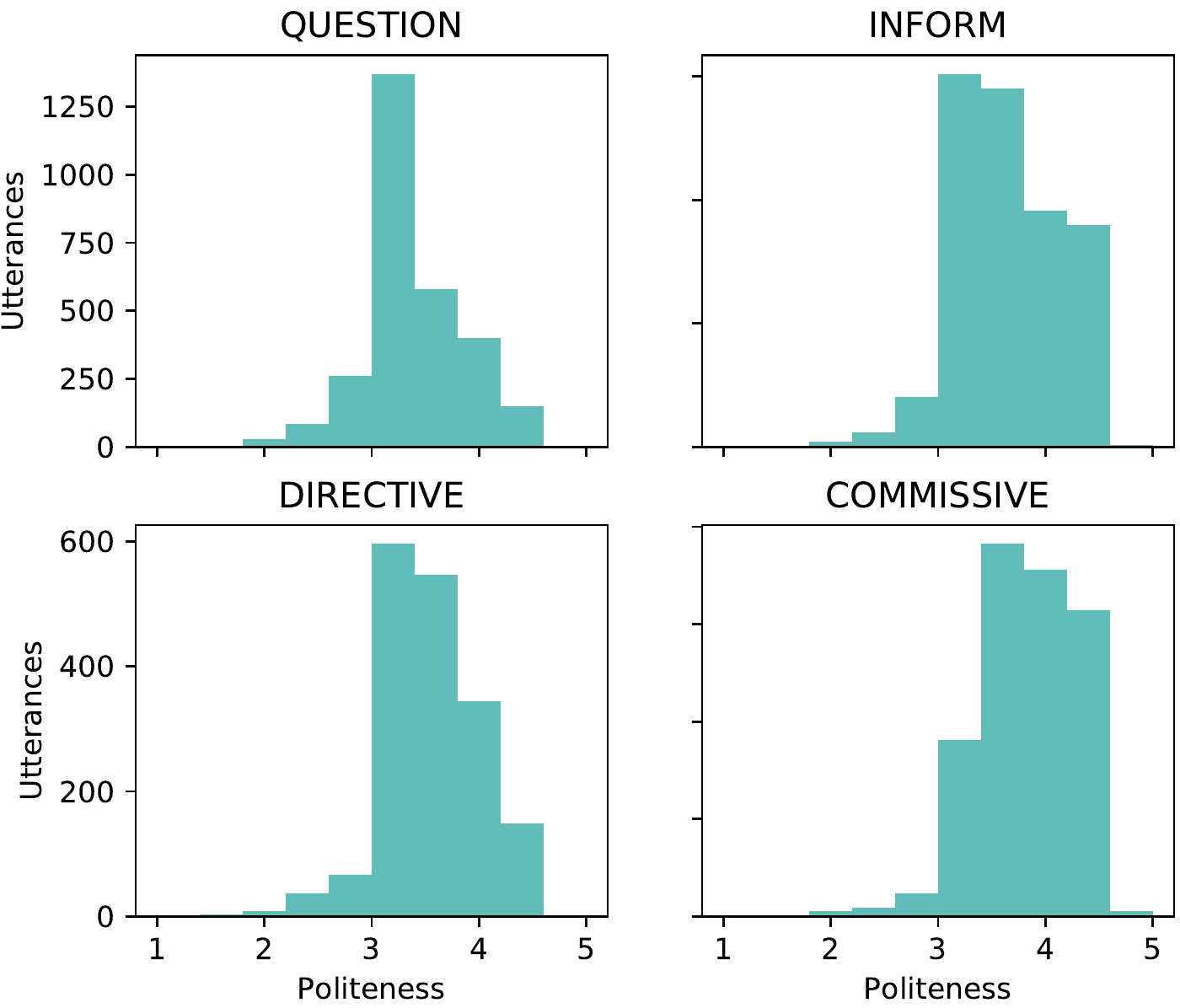} 
\caption{Politeness histograms for the dialogue acts in the DailyDialog dataset shows that the utterances in Question and Directive are mostly neutral whereas in Inform and Commissive are mostly polite}
\label{pol_histo_da}
\end{center}
\end{figure}

Finally, we sort the utterances for their respective dialogue acts according to the politeness values and plot the histograms as shown in Figure~\ref{pol_histo_da}.
We can notice that most of the utterances are neutral or polite on the politeness scale for all the dialogue acts.
However, we observe that the Inform and Commissive dialogue act classes contain many polite utterances compared to Question and Directive dialogue acts.
Interestingly, we discover that speakers in the dataset often tend to use polite utterances when answering, thanking, and agreeing (and the utterances that come under Inform and Commissive), whereas while asking, guiding, directing, ordering (under Question and Directive), most of the utterances are primarily neutral and polite.

\vspace{0.2cm}

\begin{figure}[t!]
\begin{center}
\includegraphics[width=0.5\textwidth]{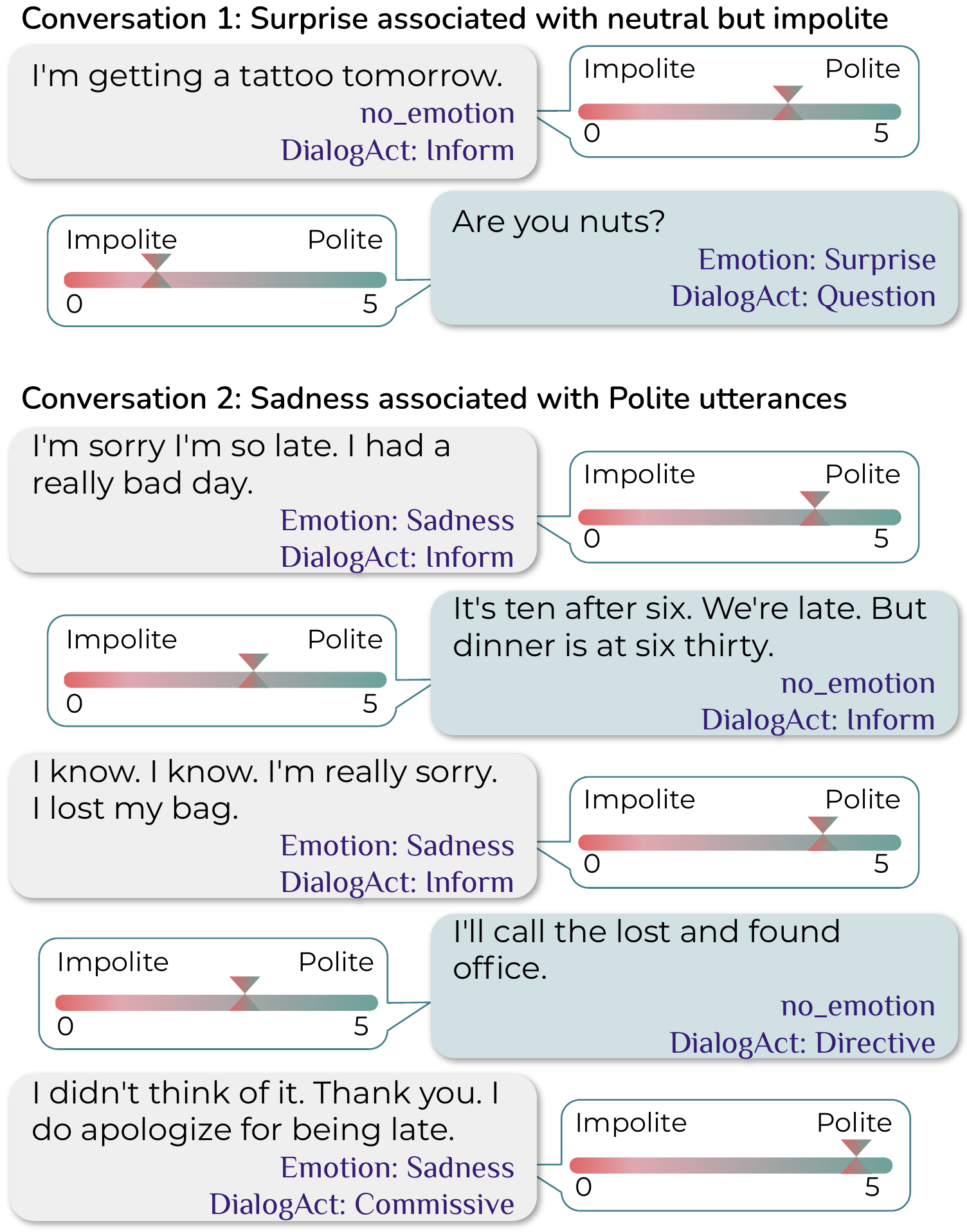} 
\caption{Polite Emotional Dialogue Act labels from the DailyDialog dataset focused on Surprise and Sadness emotion utterances, the utterance with the emotion Surprise shows an immediate reaction which becomes impolite,
interestingly most of the utterances with the emotion Sadness are polite}
\label{emotion_pol_impol_dial}
\end{center}
\end{figure}

Figure~\ref{emotion_pol_impol_dial} provides two conversation examples containing Surprise and Sadness emotion utterances.
We notice that the utterances with the emotion Sadness are primarily polite, which can be elicited from Figure \ref{pol_histo_emo}.
We also notice that most of the utterances with the Inform and Commissive dialogue acts are polite, as elicited in Figure~\ref{pol_histo_da}.
These analytical graphs provide an insightful discovery and statistical adherence of the social cues relatedness.
It also confirms the hypothesis of natural selection of politeness strategies in various emotional states and dialogue acts.

\balance

\section{Conclusion}

Recognizing politeness in conversation is an essential aspect of conversational analysis, and we discover how politeness functions in a dialogue with respect to emotional states and dialogue acts. 
Moreover, socio-linguistic feature relatedness provides an additional dimension for the behavioural analysis of conversation partners and their use in the virtual assistant and human-robot interaction fields.
This paper discovers how politeness is associated with emotions and dialogue acts in the given dataset.
Specifically, we found that the utterances are mostly polite in Happiness and Sadness emotion classes and Inform and Commissive dialogue acts.  
Similarly, the utterances are primarily neutral and impolite for the Anger and Disgust emotion classes.
We also observed that the utterances in the Question and Directive dialogue acts are mostly neutral; however, many are also polite.
We also presented examples of utterances from the data demonstrating the discovered phenomenon and made the annotated politeness data available for the research community.

\vspace{0.2cm}

A continuation of the experiment is planned to extend to other dialogue datasets to demonstrate the discovered phenomenon.
The politeness analyzer used in this experiment provided linguistically appropriate annotations for the degree of politeness.
However, in the next step, more than one analyzer could be used for a more robust detection of politeness or annotate the utterances manually. 
Future work could extend to learning the social behaviours using analyzed socio-linguistic cues with the help of deep learning techniques. 
Moreover, the majority of the utterances in the explored dataset contain no emotion class; thus, using multimodal data could improve analytical insight.

\section{Acknowledgements}
We gratefully acknowledge partial support from the German Research Foundation (DFG) under the project Crossmodal Learning (CML, Grant TRR 169).

\section{Bibliographical References}\label{reference}

\bibliographystyle{lrec2022-bib}
\bibliography{lrec2022-example}

\end{document}